\documentclass{article}

\usepackage[preprint]{corl_2026} % Uncomment for pre-prints (e.g., arxiv); This is like ``final'', but will remove the CORL footnote.

\usepackage{hyperref}
\usepackage{url}
\usepackage{amsmath} % assumes amsmath package installed
\usepackage{amssymb}  % assumes amsmath package installed
\usepackage{multicol}
\usepackage{float}
\usepackage{graphicx}
\usepackage{amsfonts}
\usepackage{booktabs}
\usepackage{array}
\usepackage{bbm}
\usepackage{ulem}
\usepackage{graphics, wrapfig}
\usepackage{subcaption}
\usepackage{caption}
\usepackage{stackengine}
\usepackage{tikz}
\usetikzlibrary{bayesnet}
\usepackage{dsfont}
\usepackage{xcolor}
\usepackage{algorithm}
\usepackage{algpseudocode}
\usepackage{multirow}
\usepackage[utf8]{inputenc}

\usepackage{tabularx} % Essential for fixing table width and text wrapping
\title{Spatio-Temporal Retrieval-based Priors for Adaptive Computational Teaching in Driving}

\usepackage{authblk}

\author{Deepak Gopinath${}^{\star}$}
\author{Xiongyi Cui${}^{\star}$}
\author{Jonathan DeCastro}
\author{Avinash Balachandran}
\author{Guy Rosman}
\affil{Toyota Research Institute, Cambridge, MA, USA, \texttt{first.last@tri.global}\\ ${}^{\star}$~Contributed equally}

\newcolumntype{Y}{>{\centering\arraybackslash}X}

\begin{document}
\maketitle

%===============================================================================

\begin{abstract}
     Learning-based automated coaching systems for complex motor tasks such as high-performance driving remain limited in the ability to be \textit{adaptive} by their reliance only on local, context-dependent reasoning, failing to account for the long-term temporal nature of student learning and the cumulative impact of repeated teacher–student interactions. In this paper, we propose an imitation learning based computational model for \textit{adaptive} teaching with a dedicated temporal reasoning module that can reason over the interaction history under low-data regimes. To compensate for limited amounts of interactive training data, and based on the repetitive nature of the teaching process, the model relies on a \textit{nearest neighbor retrieval and cross attention prior}, reasoning only on a narrowed-down set of semantically similar past interactions with an encoder-decoder based concurrent teaching model. We validate our approach with (i) a novel semi-synthetic closed-loop longitudinal student-teacher interaction dataset based on Waymo Open Motion Dataset and (ii) a small-scale real-world naturalistic simulator race coaching dataset. Our results reveal the consistent advantage of our adaptive teaching model with the nearest neighbor retrieval and cross-attention prior over a non-adaptive baseline as well as a suite of adaptive models that differ in their choice of priors and temporal fusion mechanisms.
\end{abstract}


%===============================================================================
\section{Introduction}
\vspace{-0.3cm}

The ability to coach is a fundamental capability of AI systems poised to assist humans in many domains---from medicine~\cite{rojas2020daisi,10.1001/jamasurg.2025.2564} and sports~\cite{yin2025flex,pashaie2024unlocking} to semi-autonomous driving~\cite{gopinath2025computational} and rehabilitation robotics~\cite{santos2021design}, among others~\cite{chen2020artificial}.
Coaching humans in embodied domains depends on several aspects such as context-dependency, task semantics, and people's temporal adaptation. Unlike in non-embodied domains such as classroom pedagogy, where interactive student-teacher data is available at a large scale~\cite{mayhew2020simultaneous,choi2020ednet}, in embodied settings, the feasibility of learning-based methods for training automated coaching systems is limited by the availability of high-quality interactive teaching data which further exacerbates reasoning about the longitudinal adaptation. Despite the complexity of the coaching task, humans are able to teach tractably through strong priors, heuristics and disentangling the influence of different temporal factors. Hence, learning-based solutions in low-data regimes likely require careful use of the available problem structure and task-specific priors. 

    We consider the setting of an automated coaching agent interacting with a student (human) on an embodied task, such as driving, in a long-term repeated interaction setting. 
    Our objective is to build a computational model that can provide contextually-relevant concurrent verbal feedback, \textit{adapt} its teaching over the course of interaction, be trained in a data-efficient manner via structured priors. 
    
    We note that the teaching process often contains repeated interactions~\cite{lyster1997corrective,mondada2018driving}, where similar context and student performance is coupled with suitable teacher actions. This observation suggests spatio-temporal retrieval as a data-efficient causal prior for longitudinal teaching adaptation, leading us to propose a model that performs local vs. global spatio-temporal reasoning using a modular architecture while leveraging a \textit{nearest-neighbor retrieval and cross-attention prior} to compensate for  data limitations.
    We validate our approach with experiments using both (i) a novel semi-synthetic closed-loop longitudinal student-teacher interaction dataset based on Waymo Open Motion Dataset (WOMD)~\cite{ettinger2021large} and (ii) a real-world naturalistic simulator racing coaching dataset~\cite{simcoach}.
    
    \textbf{Contributions}:
    \vspace{-0.2cm}
    \begin{itemize}
        \item We introduce a modular imitation learning framework for adaptive coaching in driving that relies on a nearest neighbor retrieval together with a cross-attention based fusion mechanism to enable data-efficient training. 
        \item We introduce a novel semi-synthetic dataset, \textsc{WayCoach} based on Waymo Open Motion Dataset that simulates closed-loop longitudinal teacher-student interaction.
        \item We demonstrate the usefulness and efficacy of the nearest neighbor retrieval and cross-attention based prior along different axes through extensive experiments on \textsc{WayCoach} and a publicly available small-scale naturalistic racing simulator based coaching dataset, \textsc{SimCoachCorpus}~\cite{simcoach}. 
    \end{itemize}
%===============================================================================
\vspace{-0.15cm}
\section{Related Work}
\vspace{-0.3cm}
    
    Our suggested approach relates to several topics of research from different communities.

    The topic of multiscale temporal context in imitation learning and behavior cloning has been studied, both in terms of multiscale architectures~\cite{le2018hierarchical,liu2024multi,fox2019hierarchical}, as well as more general approaches, such as decision transformers~\cite{chen2021decision}
    
    Within AI in education, the usage of multiscale temporal structure has been used for knowledge tracing~\cite{chen2020artificial,corbett1994knowledge,piech2015deep,zhang2017dynamic}, as well as schedule planning~\cite{pavlik2008using} and estimation of student reactions~\cite{calvo2010affect,jeong2008using} and memory and retention~\cite{settles2016trainable,balakrishnan2013predicting}.
    However, the majority of works in knowledge tracing inspected classroom and education settings, where data is plentiful and observable, lending itself to general temporal modeling approaches (e.g. LSTMs~\cite{piech2015deep} or transformers~\cite{pu2020deep}).
    % \guy{need to verify -- are there  embodied knowledge tracing works}
    
    Motor learning/teaching approaches traditionally either optimally schedule exercises~\cite{Srivastava2022-mg} or provide corrections to optimize student performance~\cite{Srivastava2023-ix}. Recent works in computational teaching of motor sports have explored teacher imitation with short episodic context~\cite{srivastava2025shared,gopinath2025computational}. While longitudinal motor learning datasets have been a bottleneck for AI, this gap has been recently reduced~\cite{simcoach}.
    
    More broadly, several machine learning research threads have looked at handling of long temporal sequences - recurrent~\cite{chung2017hierarchicalmultiscalerecurrentneural}, convolutional~\cite{yu2015multi}, attention-based~\cite{DBLP:journals/corr/abs-1901-02860}, and memory~\cite{DBLP:journals/corr/GravesWD14} or retrieval-~\cite{lewis2020retrieval,pari2021surprising} based approaches to name a few. However, these often assume a large amount of data, which is often not available for motor learning~\cite{simcoach}, and as we show in this paper, not strictly required. A closely related research direction is that of sparse temporal modeling where the challenge is to learn good model from relatively large but incomplete dataset~\cite{deng2024learning}.

    Finally, our approach is an instance of an embodied teaching policy. Approaches for teaching in embodied fields have been considered with emphasis on both decision-making~\cite{rojas2020daisi,gopinath2025computational,yu2023coach}, engagement estimation~\cite{shlomov2023ongoing}, and overall interactive system design~\cite{fuchino2022t2snaker}, among other aspects.
    Specifically, estimation of skill in embodied teaching scenarios has been considered in fields such as music~\cite{ziegenbein2022monitoring}, surgery~\cite{forestier2018surgical,kamboj2026skill}, and driving~\cite{ropelato2018adaptive}. However, traditionally, these approaches heavily rely on specific features to estimate skill, which is then used to condition the teaching process.

%===============================================================================
\vspace{-0.15cm}
\section{Technical Approach}
\vspace{-0.3cm}

    In this section, we present our problem formulation, the mathematical notation, and details about our model architecture.
    We aim to learn a computational model for teaching in embodied domains that can (i) imitate low-level teaching actions conditioned on local context and behavior as well as (ii) \textit{adapt} how it teaches over multiple interactions with the student under low-data regimes. 
    
    \subsection{Problem Formulation}
    
    We adopt and augment the notational conventions introduced in~\cite{gopinath2025computational}. Let $s^t \in \mathcal{S}$ denote the human-driven ego car states, $a_H^t$ denote the human's control actions (steering/throttle, etc.), and $a_C^t$ denote the coach's instruction at the time $t$. Let $\mathcal{T}$ be a dataset of trajectories sampled from a teacher-student interaction session, where $\tau \in \mathcal{T}$ denotes a trajectory such that $\tau = \left\{(s^t, a_H^t, a_C^t)\right\}_{t=0}^{T}$. 
    We denote the global map as $\mathcal{M}_{global}$ and the set of \textit{localized} maps for each trajectory $\tau$ with respect to the ego car pose at $t=0$ as $\mathcal{M}_{local}$. The tuple $(\tau, m_{local}) \in \mathcal{T} \times \mathcal{M}_{local}$ as a \textit{scenario}, denoted as $q \in \mathcal{Q}$. We also distinguish between $q_{curr} \in \mathcal{Q}_{curr}$, the current scenario of interest in which teaching happens and $q_{past} \in \mathcal{Q}_{past}$, which is any scenario in which teaching already happened \textit{before} the current scenario $q_{curr}$. Additionally, let $\mathcal{H} = [{q_{past}^1, \dots, q_{past}^H}$] denote a set of past scenarios with respect to $q_{curr}$. These past scenarios can be short trajectory snippets (of the order of seconds) or temporally extended behavior lasting minutes (e.g., an entire lap of driving in the racing context). 

    We define an \textit{adaptive} computational teaching policy as mapping $\Pi: (\mathcal{Q}_{curr} \times \mathcal{H}) \rightarrow \mathcal{A}$, where $\mathcal{A}$ is a set of teacher action categories that are output on the current scenario $q_{curr}$. In the context of driving, the action space $\mathcal{A}$ includes the set of all possible utterances by the instructor, e.g., $[\texttt{brake}, \texttt{accelerate}, \texttt{stay left}, \texttt{stay right}, \texttt{turn}]$, along with a \texttt{no-op} action. When $\mathcal{H} = \emptyset$, the teaching policy becomes \textit{non-adaptive} and depends only on the current scenario $q_{curr}$. 
    
    \subsection{Model Components}
\begin{figure*}
        \centering
        \includegraphics[width=1.0\linewidth]{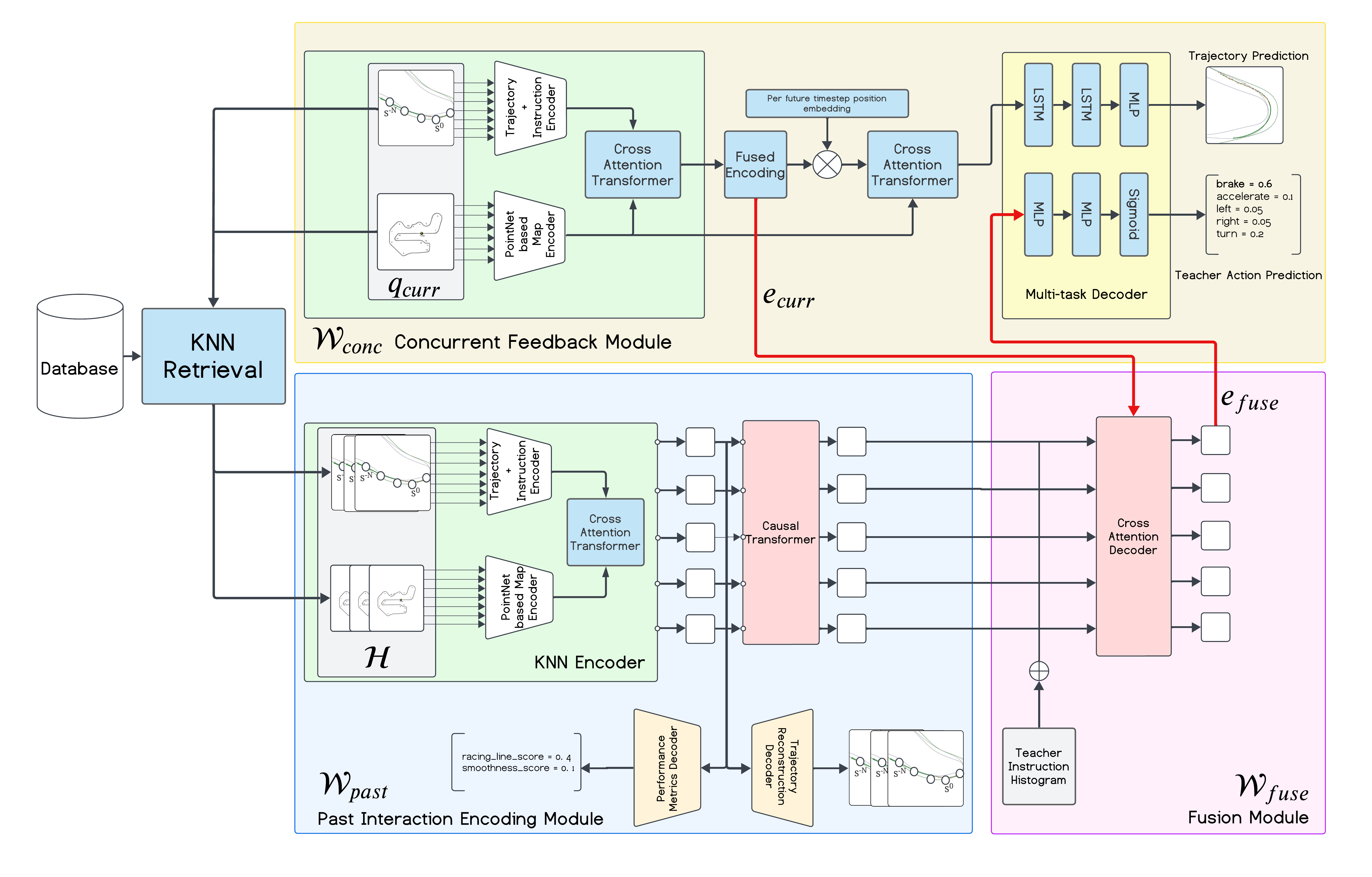}
        \caption{Overall model architecture diagram. A multi-task concurrent feedback teacher imitation module is augmented by a history fusion module that encodes a set of $K$ nearest teaching scenarios from the student-teacher interaction history.}
        \label{fig:model_diagram}
    \end{figure*}

    In this section, we present the details of the three modules that constitute the adaptive computational teaching model shown in Figure~\ref{fig:model_diagram}.
    We adopt a multi-task approach to imitation learning (IL) due to its advantages in learning good representations when data is limited~\cite{gopinath2025computational}. We consider multiple self-supervised tasks such as trajectory prediction and reconstruction as well as metrics and skill prediction as our auxiliary tasks. 
    Our model consists of three separate modules: (i) an encoder-decoder based \textit{concurrent teaching module}, $\mathcal{W}_{conc}$, that operates on local spatio-temporal context, (ii) a \textit{past interaction encoding module}, $\mathcal{W}_{past}$, that operates on selective episodes retrieved via a nearest neighbor prior from the student-teacher interaction history and lastly (iii) \textit{a fusion module}, $\mathcal{W}_{fuse}$, that fuses the encoded information from $\mathcal{W}_{past}$ into $\mathcal{W}_{conc}$ to aid adaptation. 
   
    \subsubsection{Concurrent Teaching Module ($\mathcal{W}_{conc}$)}

    \textbf{Encoder}: The encoder module for $\mathcal{W}_{conc}$ consists of two parallel streams: a) an MLP+Transformer based trajectory encoder that encodes past trajectory $\mathcal{\tau}_{1:N}$, with $N < T$ and b) a PointNet~\cite{qi2017pointnet} based map encoder consisting of interleaved layers of MLP and transformer based encoders that encodes the local map $m_{local}$. We use a set of polylines to represent the left, center and right edges of each road lane to represent $m_{local}$. Each polyline consists of sequence of nodes ordered according to a prespecified lane direction and each node is a 2D vector of normalized position. The trajectory and map encodings are combined using a cross attention transformer to produce the encoded state, $e_{curr}$ which then gets fused with encodings that capture the temporal history generated from $\mathcal{W}_{past}$.

    \textbf{Decoder}: The output decoders for the teacher action prediction consists of a series of MLP-based transformations with a final sigmoid activation layer and operates on the encodings, $e_{fuse}$ from the fusion module $\mathcal{W}_{fuse}$ in the case of a full adaptive model or from $\mathcal{W}_{conc}$ for the non-adaptive model. For the auxiliary task of trajectory prediction, we use an LSTM-based decoder with anchors~\cite{gu2021densetnt}.

    \subsubsection{Past Interaction Encoding Module ($\mathcal{W}_{past}$)}

    \textbf{Encoder}: $\mathcal{W}_{past}$'s encoder is akin to that of $\mathcal{W}_{conc}$ and operates in parallel on the scenarios selected from $\mathcal{H}$. The generated encodings are augmented with time embeddings (for example, lap number or timesteps) and then processed via a causal transformer to generate representations that aggregate information through time. 
    Since we have access to the full history, unlike $\mathcal{W}_{conc}$, we rely on trajectory reconstruction and metrics prediction as our self-supervised tasks. The final encodings corresponding to each of the scenarios in $\mathcal{H}$ are then fed as input to the fusion module $\mathcal{W}_{fuse}$.

    \textbf{Decoder}: Similar to $\mathcal{W}_{conc}$'s decoders, $\mathcal{W}_{past}$ uses an LSTM-based decoder to do trajectory reconstruction and an MLP-based decoder for metrics prediction. 
    
    \subsubsection{Fusion Module ($\mathcal{W}_{fuse}$)}
    $\mathcal{W}_{fuse}$ augments the representation learned within $\mathcal{W}_{conc}$ with information aggregated by $\mathcal{W}_{past}$ from a subset of scenarios $\mathcal{P}$ in $\mathcal{H}$. We use a cross-attention based transformer decoder module to facilitate the fusion between $e_{curr}$ and $\mathcal{W}_{past}$'s encodings. After the transformer processes the inputs the token encoding with maximum magnitude is added to $e_{conc}$ to form an \textit{past-interaction aware adaptive} representation, $e_{fuse}$, for generating the teaching action in $q_{curr}$. If $\mathcal{W}_{fuse}$ is a null operation, then no information from $\mathcal{W}_{past}$ flows into $\mathcal{W}_{conc}$ and the overall model becomes \textit{non-adaptive} and unaware of how the past scenarios influence teaching in $q_{curr}$.
    
    \subsection{Nearest Neighbor Prior}
    The most general instantiation of $\mathcal{H}$ is to utilize the entire driving history (entire laps up until the current laps or a full memory bank of all past driving scenarios) up until $q_{curr}$ to inform teacher adaptation. 
    Although an attention-based scheme provides the correct inductive bias, the data requirements to recover the underlying temporal correlations in an end-to-end fashion are quite high. In embodied interactive domains, data collection is expensive, and large-scale datasets can be prohibitive. Our key insight is to 
    provide a data prior via a $K$-nearest neighbor mechanism that utilizes spatio-temporal features (such as global position and velocity) to retrieve relevant scenarios to improve model performance. The choice of features depends on the task and is critical for extracting the right set of priors. During training, we ensure that the retrieved samples are temporally ordered to ensure proper causal propagation of information. 
    
    \subsection{Loss Functions}
    We adopt a multi-task IL paradigm for $\mathcal{W}_{conc}$ and $\mathcal{W}_{past}$ due to its demonstrated advantages~\cite{gopinath2025computational}. 
    
    \textbf{Concurrent Feedback Module Loss $\mathcal{L}_{conc}$}: 
    The main component of  $\mathcal{L}_{conc}$ is the teacher action prediction loss, $\mathcal{L}_{teacher}$. We cast teacher action prediction as \textbf{binary multi-label prediction} in which the multi-label target represents the presence or absence of a teaching action category in the future time window.
    $\mathcal{L}_{teacher}$ is a weighted Binary Cross Entropy (wBCE) with class weights corresponding to the ratio of negative to positive samples in each class computed over the training set. 
    For the auxiliary task of trajectory prediction we use an MoN version of Average Displacement Error as the loss, $\mathcal{L}_{traj-pred}$. 
    $\mathcal{L}_{conc}$ is then a weighted sum of $\mathcal{L}_{teacher}$ and $\mathcal{L}_{traj-pred}$.
    
    \textbf{Past Encoder Loss $\mathcal{L}_{past}$}: The self-supervised task of trajectory reconstruction also utilizes a MoN version of Average Displacement Error ($\mathcal{L}_{traj-reco}$) and metrics prediction relies on an MSE loss, ($\mathcal{L}_{metrics}$). $\mathcal{L}_{past}$ is then a weighted sum of $\mathcal{L}_{traj-reco}$ and $\mathcal{L}_{metrics}$.
    
    The overall loss $\mathcal{L}$ is then given by $
        \mathcal{L} = \mathcal{L}_{conc} + \mathcal{L}_{past}$. 
    Note that, as all the modules are co-trained, gradients from teacher action decoding flow back into $\mathcal{W}_{past}$ and therefore $\mathcal{L}_{conc}$ also affects $\mathcal{W}_{past}$.

%===============================================================================
\vspace{-0.15cm}
\section{Experimental Results}
\vspace{-0.3cm}

\label{sec:result}
In this section, we describe our datasets, evaluation metrics and experimental results on the impact of the nearest neighbor retrieval and cross attention prior under different training data regimes.

% \subsection{Datasets}
\subsection{\textsc{WayCoach}: Semi-Synthetic Longitudinal Teaching Dataset}
To systematically investigate the impact of the prior under data constraints we curate a new semi-synthetic longitudinal teaching dataset, \textsc{WayCoach}, using naturalistic driving scenarios from the Waymo Open Motion Dataset (WOMD)~\cite{ettinger2021large}. We build off the approach presented in~\cite{gopinath2025computational} with few key improvements that adds to the realism. First, we introduce longitudinal student performance improvement via temporal skill modeling. Second, we model the closed-loop effect of a teaching action on performance improvement. The entire dataset consists of multiple \textit{sequences} of length $T$ of scenarios sampled from WOMD. Each sequence represents a single student's learning journey which is modeled as the time evolution of a two-dimensional latent skill vector $\boldsymbol{z}^t = (\alpha^t, \beta^t)$ that characterizes their driving (how aggressive ($\beta$) vs. conservative ($\alpha$)) at learning timestep $t$. Additionally, the temporal evolution of $\boldsymbol{z}^t$ is governed by the following parameters: 

(i) the base learning rate, $\eta$ (ii) the overall tendency, $b_{aggr}$, to be aggressive/conservative at $t=0$ and (iii) an \textit{influence} parameter, $\lambda$ that captures how much they get influenced by the teacher's past actions.  The discrete-time dynamical system that governs the time-evolution of $\boldsymbol{z}^t$ is given by 
\begin{equation}
    \boldsymbol{z}^{t+1} = \boldsymbol{z}^t + \eta(\boldsymbol{z}^{final} - \boldsymbol{z}^t) - \lambda f(\{a_C^h\}_{q_h \in \mathcal{P}}) + \epsilon
\end{equation}
where $\epsilon \sim\mathcal{N}(0, \sigma^2)$ is per timestep IID Gaussian noise, $f(\cdot)$ is the teacher feedback function operating on scenarios from $\mathcal{P} \subseteq \mathcal{H}$ and $\boldsymbol{z}^0$ is initialized as $[z^0\cdot b_{aggr}, z^0\cdot (1-b_{aggr})]^{T}$ with $z^0 \sim (0.95, 1.0)$ and $\boldsymbol{z}^{final}$ is the asymptotic limit for $\boldsymbol{z}$ (highest skill possible). 
The skill progression happens in such a way that as $t\to T$, the sampled scenarios will primarily be from the \textit{neutral} group. The interpretation is that at $t$=$0$, a student is unskilled and could be highly aggressive or conservative (as they are figuring out how to drive); as time passes, due to self-learning and the effect of teaching, they become less aggressive and conservative and begin to drive stably.
We focus on interactive maneuvers such as yields and merges. The parameters $b_{aggr} \in [0.1, 0.5, 0.9]$ and $\eta \in$ $[slow$$\sim$$(0.2, 0.5)$, $medium$$\sim$$(0.8, 1.5)]$ together result in six distinct student types. For each student type, we sample 67 students (15 sequences per student) resulting in a total of $C$=$6030$ sequences each of length $T$=$20$ timesteps. Each sequence represents a student's learning journey. 
A teaching action $a_C^h$ from a predefined set of teaching actions $[\texttt{no\_op}, \texttt{slow\_down}, \texttt{speed\_up}]$ is assigned to each scenario $q_h$ in the sequence. We model two types of synthetic teachers; (i) a \textit{nonadaptive} teacher whose teaching action only depends on the scenario label and an (ii) \textit{adaptive} teacher whose teaching action takes into consideration a skill estimate (computed as the proportion of \textit{conservative} vs. \textit{aggressive} scenarios in $\mathcal{P} \subseteq \mathcal{H}$, is the set of nearest neighbors to each scenario with $|\mathcal{P}|=4$) when generating the teaching label. More details about the construction of the dataset are in the Supplementary material. 
\vspace{-0.2cm}
\subsection{\textsc{SimCoachCorpus}: Naturalistic Simulator Racing Instruction Dataset}
\vspace{-0.2cm}
To study the efficacy of the nearest neighbor retrieval and cross-attention prior in a naturalistic real-life coaching domain (high performance driving education) we use a publicly available dataset, \textsc{SimCoachCorpus}~\cite{simcoach}. We only use data from the \textbf{coached} condition, notably a small dataset with 15 students; therefore is a perfect testbed for our research questions. 
In the \textbf{coached} condition, each student undergoes training on an average of 22 laps with demonstrated adaptation of coaching. We set the teacher action target to be 8D (we ignore steering and turn categories due to the very low number of positive labels), and each training snippet is 10s, with the first 5s treated as the input and the last 5s as the prediction target. The retrieved nearest neighbor samples are also 10s long. 
\vspace{-0.2cm}
\subsection{Evaluation Metrics}
\vspace{-0.2cm}
We report standard multiclass (weighted $F_1$-score) and multilabel (weighted $F_1$-score and Hamming distance) classification metrics on the teacher action prediction for \textsc{WayCoach} and \textsc{SimCoachCorpus} respectively. Results reported on \textsc{WayCoach} are averaged over five random seeds; experiments with \textsc{SimCoachCorpus} uses a 15-fold cross validation with four seeds per fold. The result tables report the average of the max scores over all seeds.

\vspace{-0.2cm}
\subsection{How does a K-nearest neighbor retrieval prior help?}\label{ssec:how_does_KNN_help}
\vspace{-0.2cm}
\begin{table}[t]  % [b] places at bottom of page; use [btp] to give LaTeX more options
    \centering
    
    % Table 1
    \caption{Average of Maximum Weighted $F_1$-score ($\uparrow$) for \textsc{WayCoach} experiments on \textit{adaptive} teaching. For small data regime (top row) \textbf{KNN-CrossAttn} outperforms \textbf{Full-CrossAttn}.} 
     \label{tbl:table_1}
     {\renewcommand\tabularxcolumn[1]{m{#1}}
     \begin{tabularx}{\textwidth}{|l|Y|Y|Y|Y|}
       
        \hline
        \textbf{Dataset-Size} & \textbf{NonAdaptive} & \textbf{Random-CrossAttn} & \textbf{Full-CrossAttn} & \textbf{KNN-CrossAttn (Ours)} \\ \hline
        60 & \small \mbox{$0.401 \pm 0.042$} & \small \mbox{$0.534 \pm 0.041$} & \small \mbox{$0.574 \pm 0.029$} & \small \mbox{$\textbf{0.614}\pm \textbf{0.043}$} \\ \hline
        600 & \small \mbox{$0.476 \pm 0.006$} & \small \mbox{$0.582 \pm 0.021$} & \small \mbox{$\textbf{0.891}\pm \textbf{0.018}$} & \small \mbox{$0.701 \pm 0.016$}  \\ \hline
        6000 & \small \mbox{$0.560 \pm 0.017$} & \small \mbox{$0.670 \pm 0.009$} & \small \mbox{$\textbf{0.980}\pm \textbf{0.002}$} & \small \mbox{$0.835 \pm 0.016$}   \\ \hline
    \end{tabularx}
    }
    \vspace{1em}
    
    % Table 2
    \caption{Average of Maximum Weighted $F_1$-score ($\uparrow$) and Hamming distance ($\downarrow$) on the naturalistic \textsc{SimCoachCorpus} dataset. Results averaged over 60 seeds (4 seeds per fold for 15-fold cross validation) with $K$=$7$.}
    \label{tbl:table_2}
    {\renewcommand\tabularxcolumn[1]{m{#1}}
    \begin{tabularx}{\textwidth}{|l|Y|Y|Y|Y|}
     
        \hline
        % \multicolumn{1}{|c|}{\textbf{Metric}}  
        & \textbf{NonAdaptive} & \textbf{Random-CrossAttn} & \textbf{Full-CrossAttn} & \textbf{KNN-CrossAttn (Ours)} \\ \hline
        \small Weighted $F_1$-score $\uparrow$ &\small \mbox{$0.701 \pm 0.031$} & \small \mbox{$0.711 \pm 0.031$}  & \small \mbox{$0.723 \pm 0.029$} & \small \mbox{$\textbf{0.728} \pm \textbf{0.026}$} \\ \hline
        \small Weighted Hamming Distance $\downarrow$ &\small \mbox{$0.303 \pm 0.026$} & \small \mbox{$0.303 \pm 0.030$} & \small \mbox{$0.286 \pm 0.025$} & \small \mbox{$\textbf{0.286} \pm \textbf{0.024}$} \\ \hline
    \end{tabularx}
    }
    
    \vspace{1em}
    
    % Table 3
    \caption{Average of Maximum Weighted $F_1$-score ($\uparrow$) for different fusion mechanisms with $K$=$7$. \textbf{KNNCross-Attn} exhibits better performance compared to other neural and non-neural baselines.}
    \label{tbl:table_3}
    {\renewcommand\tabularxcolumn[1]{m{#1}}
    \begin{tabularx}{\textwidth}{|l|Y|Y|Y|Y|Y|}
    
        \hline
        \multirow{2}{*}{} & \multicolumn{5}{c|}{Fusion Mechanism} \\ \cline{2-6} 
         & \textbf{1-NN} & \textbf{Naive-KNN-Mean} & \textbf{KNN-MaxPool} & \textbf{KNN-MaxPool-Residual} & \textbf{KNN-CrossAttn (Ours)} \\ \hline
        \textsc{WayCoach} & \small \mbox{$0.549 \pm 0.000$}& \small \mbox{$0.577 \pm 0.000$} & \small \mbox{$0.584 \pm 0.058$}& \small \mbox{$0.592 \pm 0.058$} & \small \mbox{$\textbf{0.614} \pm \textbf{0.043}$} \\ \hline
        \textsc{SimCoachCorpus} & \small \mbox{$0.539 \pm 0.041$} & \small \mbox{$0.663 \pm 0.026$} & \small \mbox{$0.721 \pm 0.028$} & \small \mbox{$0.720 \pm 0.027$} & \small \mbox{$\textbf{0.728} \pm \textbf{0.026}$} \\ \hline
    \end{tabularx}
    }
    
    \label{tab:entire_layout}
\end{table}

\begin{table}[t]
\caption{Average of Maximum $F_1$-score as a function of $K$ for both datasets. Model performance peaks for medium values of $K$. For \textsc{WayCoach}, the true maximum occurs when $K=|\mathcal{P}|=4$.}
\label{tbl:table_4}
\centering
\begin{tabular}{lcc}
% \label{tbl}
\hline
$K$ & \textsc{SimCoachCorpus} & \textsc{WayCoach} \\
\hline
$0$       & 0.701 $\pm$ 0.031 & 0.417 $\pm$ 0.036 \\
$1$       & 0.720 $\pm$ 0.031 & 0.571 $\pm$ 0.030 \\
$3$       & 0.726 $\pm$ 0.029 & \textbf{0.613} $\pm$ \textbf{0.047} \\
$5$       & 0.726 $\pm$ 0.028 & 0.606 $\pm$ 0.045 \\
$7$       & \textbf{0.728}$\pm$ \textbf{0.026} & 0.575 $\pm$ 0.027 \\
$9$       & 0.727 $\pm$ 0.027 & 0.587 $\pm$ 0.024 \\
\hline
\end{tabular}

\label{tab:results}
\end{table}
To isolate the effect of a KNN prior on teacher action prediction we compare our full adaptive model with a $K$-nearest neighbors plus cross-attention prior, denoted as \textbf{KNN-CrossAttn} to the following baselines; (i) \textbf{NonAdaptive}: This model uses $\mathcal{W}_{conc}$ to predict the teacher action, (ii) \textbf{Full-CrossAttn}: This version of the adaptive model utilizes \textit{all available data} up until $q_{curr}$. For experiments with \textsc{SimCoachCorpus} we only use full data from $K$ previous laps. (iii) \textbf{Random-CrossAttn}: This baseline uses randomly sampled $K$ scenarios from $\mathcal{H}$ as an uninformed prior. 

\begin{wrapfigure}{r}{0.43\textwidth}
    \centering
    \includegraphics[width=\linewidth]{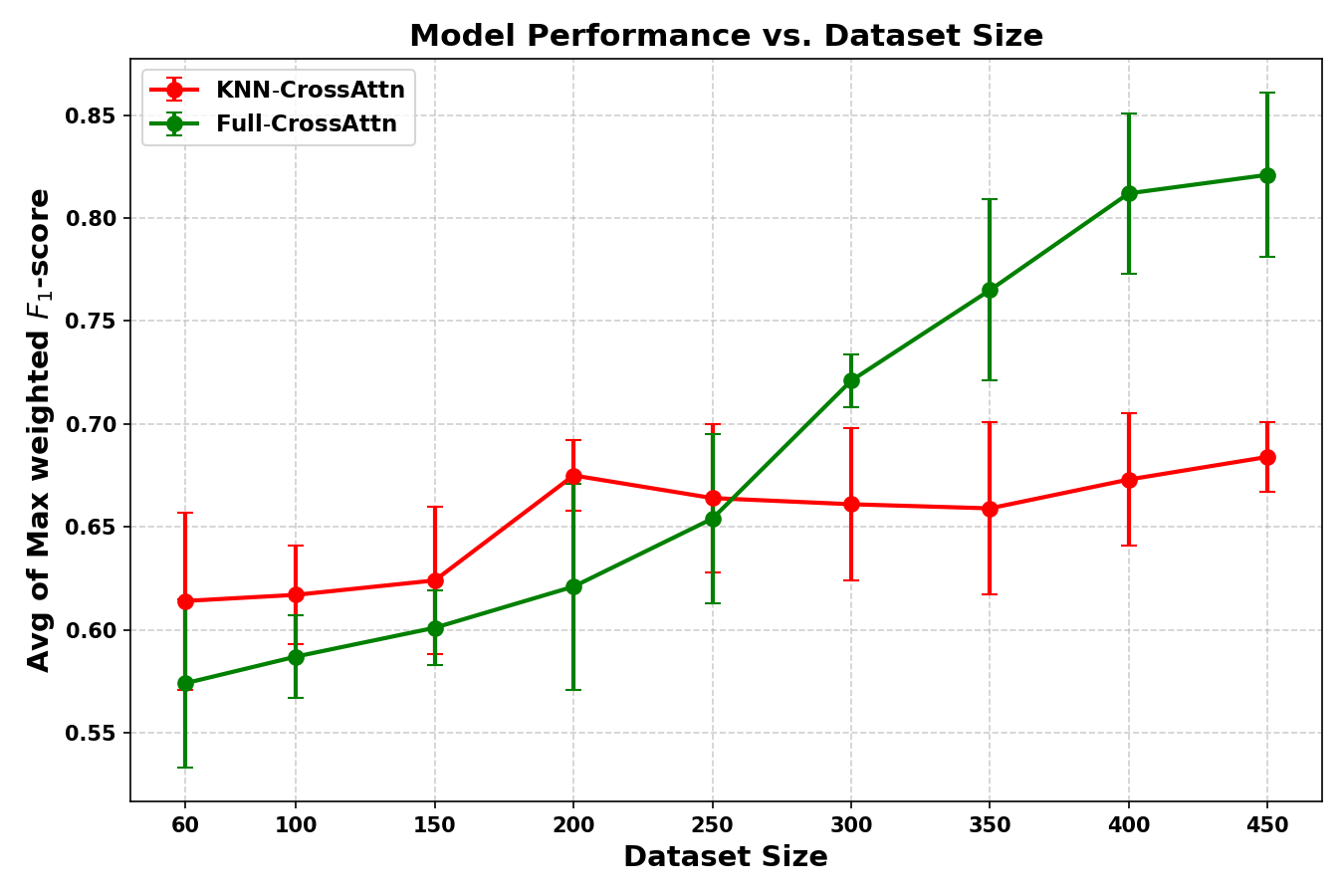}
    \caption{Model performance as a function of dataset size. \textbf{KNN-CrossAttn} exhibits a gradual degradation compared to \textbf{Full-CrossAttn} for smaller amounts of data.}
    \label{fig:model_performance_vs_dataset_size}
    \vspace{-0.5cm}
\end{wrapfigure}

\textbf{Results on \textsc{WayCoach}}:
Table \ref{tbl:table_1} reports results when trained on the adaptive teacher dataset. In the large data regimes (bottom row) \textbf{Full-CrossAttn} condition results in a very high validation performance. This is because, for longer sequences, the transformer's attention mechanism effectively treats the encoded teaching actions at all previous time-steps as \textit{in-context examples}~\cite{garg2022can} and is able to recover the underlying generative mechanism that produces the teaching action. In contrast, although the \textbf{KNN-CrossAttn} model sees $\mathcal{P}$ as its input, there is not enough information in its context to recover the generative mechanism and hence does not generalize right away. However, in low-data regimes (top row of \ref{tbl:table_1}), despite overall performance being lower, \textbf{KNN-CrossAttn} outperforms \textbf{Full-CrossAttn}. We observe in Figure \ref{fig:model_performance_vs_dataset_size} that \textbf{KNN-CrossAttn} model degrades more gradually compared to \textbf{Full-CrossAttn} which exhibits a sudden drop in performance for dataset sizes less than 400.  \textbf{NonAdaptive} model (leftmost column of Table~\ref{tbl:table_1}) performs the worst as it does not have access to any relevant information regarding teacher adaptation from $\mathcal{H}$. Although uninformed, in the \textbf{Random-CrossAttn} condition access to additional scenarios and teaching labels effectively acts as a data based prior and marginally improves performance compared to \textbf{NonAdaptive}. All these results taken together indicate that in low-data regimes a nearest neighbor plus cross-attention prior is helpful.
\textbf{Results on \textsc{SimCoachCorpus}}: Table \ref{tbl:table_2} shows that in the naturalistic domain \textbf{KNN-CrossAttn} is marginally better than all other baselines resulting in a 2.7$\%$ gain over the \textbf{NonAdaptive} model. For \textbf{KNN-CrossAttn}, the nearest neighbor computation uses a 4D feature set comprising of global position on the track, velocity and deviation from racing line. Notably, even though \textbf{Full-CrossAttn} utilized \textit{complete} lap-level trajectory data from previous $K$ laps (effectively $\sim$6x amounts of past context compared to \textbf{KNN-CrossAttn}), training was approximately $\sim$16x slower, exhibited higher overfitting trends and variance, and resulted in lower overall averaged maximum performance highlighting the usefulness of a global location based prior in achieving comparable performance with higher training efficiency. Similar to the trends observed in \textsc{WayCoach}, \textbf{Random-CrossAttn} shows better performance \textbf{NonAdaptive} condition, but does not match \textbf{KNN-CrossAttn} and \textbf{Full-CrossAttn}.
We note that while LSTM-based approaches are possible, straightforward implementation of them would fail due to the need to attend to temporally distant exemplars. An LSTM-based approach we incorporated over the full lap information resulted in consistently inferior performance compared to a transformer based backbone in $\mathcal{W}_{past}$.

\subsection{What choice of $\mathcal{W}_{fuse}$ is most effective?}
\vspace{-0.1cm}
Given that nearest neighbor prior is effective at low data regimes, we investigate how different fusion mechanisms affect model performance. We compare \textbf{KNN-CrossAttn} to the following non-neural and neural baselines. (i) \textbf{1-NN}: The multilabel teaching action of the nearest neighbor is used as the prediction for $q_{curr}$. (ii) \textbf{Naive-KNN-Mean}: The multilabel teaching actions from $K$ nearest neighbors are averaged, normalized and then thresholded to form the prediction for $q_{curr}$. (iii) \textbf{KNN-MaxPool}: A neural baseline in which the fusion mechanism is a simple maxpool operation on the KNN encodings followed by concatenation with $e_{curr}$ to form $e_{fuse}$. (iv) \textbf{KNN-MaxPool-Residual}: Another variation in which the maxpooled KNN encodings and $e_{curr}$ are processed via a lightweight MLP-based residual network to form $e_{fuse}$ which is then added back into $e_{curr}$.

\textbf{Results on \textsc{WayCoach}}:
 For \textsc{WayCoach} (Table~\ref{tbl:table_3} top), we see that the cross-attention based mechanism outperforms all other baselines. Cross attention provides the model with enough capacity to attend to key features from past scenarios and helps to learn the correlation between map context, vehicle behavior and teaching action associated with the scenario. In contrast, maxpool based operations are done independent of how much $q_{curr}$ \textit{aligns} with the nearest neighbor encodings. The non-neural baselines do not explicitly reason about the past, therefore exhibit drastically lower performance compared to the neural models. 

\textbf{Results on \textsc{SimCoachCorpus}}: For \textsc{SimCoachCorpus}, \textbf{KNNCross-Attn} outperforms all other baselines (Table~\ref{tbl:table_3} bottom). In particular, neural mechanisms drastically outperform the non-neural baselines ($\sim$6.4$\%$ improvement of \textbf{KNN-CrossAttn} over \textbf{Naive-KNN-Mean}), indicating that explicit reasoning about the past trajectories benefits adaptive teaching. Compared to a simple maxpooling, the cross-attention mechanism in \textbf{KNNCross-Attn} model attends to different nearest neighbor encodings and teases out useful representations for improving the teacher action prediction. 

\vspace{-0.05cm}
\subsection{How does the choice of $K$ affect model \vspace{-0.1cm}
performance?}
 Table \ref{tbl:table_4} shows that how model performance improves as a function of $K$. For \textsc{SimCoachCorpus}, we observe that even with $K$=$1$ there is a considerable gain in performance. Since the dataset consists of mostly novice drivers, it is likely that the instructions issued in a single nearest neighbor is still highly relevant to the current scenario, corroborated by a relatively high score for \textbf{1-NN} in Table \ref{tbl:table_3}. The peak value occurs at $K$=$7$ possibly reflecting different retrieval effectiveness of the KNN similarity metrics and the complexity of the task in that dataset. We observe similar trends with \textsc{WayCoach}, where the maximum value occurs at $K=|\mathcal{P}|=4$ (reported in Table \ref{tbl:table_1}, top row). As $K$ deviates from the peak value, the performance exhibits a slight inverted U-shaped curve.

\vspace{-0.15cm}
\section{Limitations}
\vspace{-0.3cm}
First, for KNN to work effectively with temporally \textit{sparse} past interactions in the teaching process, the choice of features matter. In race driving, the repeated structure of the task provides a natural set of spatial features to operate with in \textsc{SimCoachCorpus}. The \textsc{WayCoach} experiments are by design constructed with clear relevant features in mind.  For other domains, priors need to be chosen carefully, e.g., via domain expertise or small amount of label based supervision. Second, at test-time deployment, the model would ingest its own emissions into the input buffer, and care must be taken to reduce closed-loop and out-of-distribution effects, e.g. via priors from non-adaptive teaching policies. Third, \textsc{SimCoachCorpus} dataset also contains \textit{terminal feedback}, which is verbal feedback given at the end of each lap. Our current model fully ignores information present in the terminal feedback which could contain information on recently completed or past laps, specific areas of weaknesses, teaching plans for the subsequent lap etc. Future work will explore language modeling as an additional task and they leverage the representations corresponding to the language output as an additional side-channel information for the concurrent feedback module. 
\vspace{-0.15cm}
\section{Conclusion}
\vspace{-0.3cm}
In this paper, we introduce a framework for data-efficient multi-task imitation learning for adaptive teaching in embodied domains. We demonstrate the usefulness of a nearest neighbor retrieval and cross-attention prior through extensive experiments on a novel semi-synthetic longitudinal teaching dataset based on the Waymo Open Motion Dataset and a publicly available naturalistic simulator-based race coaching dataset. We present result on how a $K$-nearest neighbor along with a cross attention based fusion mechanism yields superior performance under data constraints when evaluated using standard multi-class and multi-label classification metrics. 
    
    % Submission to CoRL 2026 will be entirely electronic, via a web site (not email). Information about the submission process and \LaTeX{} templates are available on the conference web site at \url{https://corl.org/}. For camera ready submission, use the \texttt{final} option for the \texttt{\textbackslash usepackage} command. 

%==============================================================================

% \section{Conclusion}

% \label{sec:conclusion}

%===============================================================================

% \clearpage
% The acknowledgments are automatically included only in the final and preprint versions of the paper.
% \acknowledgments{If a paper is accepted, the final camera-ready version will (and probably should) include acknowledgments. All acknowledgments go at the end of the paper, including thanks to reviewers who gave useful comments, to colleagues who contributed to the ideas, and to funding agencies and corporate sponsors that provided financial support.}

%===============================================================================

\bibliography{corl_2026}  % .bib
\clearpage
\setcounter{section}{0}
% \section*{Appendix}

\section{Additional \textsc{WayCoach} Results}
\subsection{Effect of $a_C^h$ encoding as part of input} 
% Without prevacionencoding
We saw in Table \ref{tbl:table_1} that \textbf{Full-CrossAttn} is easily able to achieve high validation scores compared to \textbf{KNN-CrossAttn} in large data regimes. This is because, the \textbf{Full-CrossAttn} model, which utilizes \textit{entire} past history of a given scenario as the input to the model, is able to treat the input teaching actions from the previous timesteps as \textit{in-context} examples and therefore is able to recover the underlying generative rule that produced the ground truth teaching actions. To test this further, we perform an ablation in which the previous teaching action was no longer included as part of the input to the model. In Table \ref{supp_tbl:table_1_noprevaction}, we present the results from this ablation. We observe that the advantage the \textbf{Full-CrossAttn} possessed in large data regimes when previous action was part of the input has disappeared altogether confirming that the model was indeed using the input labels as in-context examples. Without previous teaching action as input, each model type needs to learn the mapping from scenarios to teaching action purely from vehicle behavior and map context, which is a harder problem without the right set of priors. And as a result, \textbf{KNN-CrossAttn}, with its nearest neighbor prior is able to outperform the other models. 
\subsection{Results on nonadaptive teacher}
In Table \ref{supp_tbl:table_2_nonadaptive} we present results on a dataset that simulates a \textit{nonadaptive} teacher. A nonadaptive teacher is that which generates the teaching action solely based on $q_{curr}$'s scenario label and therefore does not perform any kind of temporal reasoning. Since attention to past scenarios does not have an impact on model performance, \textbf{KNNCross-Attn} model is not particularly advantageous and the performance is comparable to \textbf{Random-CrossAttn}. Additionally, when previous action is encoded as part of the input, even in the small data regime (Table \ref{supp_tbl:table_2_nonadaptive}, top row) the \textbf{Full-CrossAttn} model is able to learn the deterministic mapping from scenario label to teaching action quite well. However, \textit{without} previous actions as input (Table \ref{supp_tbl:table_3_nonadaptive}), the relative advantage of the \textbf{Full-CrossAttn} model almost completely disappears and the different model types are all comparable. 
\begin{table}[!b]  % [b] places at bottom of page; use [btp] to give LaTeX more options
    \centering
    
    % Table 2
    \caption{Average of Maximum Weighted $F_1$-score ($\uparrow$) for \textsc{WayCoach} experiments on \textit{adaptive} teaching with \textbf{no previous action as input}.} 
     \label{supp_tbl:table_1_noprevaction}
     {\renewcommand\tabularxcolumn[1]{m{#1}}
     \begin{tabularx}{\textwidth}{|l|Y|Y|Y|Y|}
       
        \hline
        \textbf{Dataset-Size} & \textbf{NonAdaptive} & \textbf{Random-CrossAttn} & \textbf{Full-CrossAttn} & \textbf{KNN-CrossAttn (Ours)} \\ \hline
        60 & \small \mbox{$0.401 \pm 0.042$} & \small \mbox{$0.410 \pm 0.036$} & \small \mbox{$0.423 \pm 0.048$} & \small \mbox{$\textbf{0.448} \pm \textbf{0.055}$} \\ \hline
        600 & \small \mbox{$0.451 \pm 0.023$} & \small \mbox{$0.500 \pm 0.025$} & \small \mbox{$0.493\pm 0.030$} & \small \mbox{$\textbf{0.590} \pm \textbf{0.022}$}  \\ \hline
        6000 & \small \mbox{$0.558 \pm 0.011$} & \small \mbox{$0.619 \pm 0.021$} & \small \mbox{$0.651 \pm 0.019$} & \small \mbox{$\textbf{0.767} \pm \textbf{0.022}$}   \\ \hline
    \end{tabularx}
    }

    \label{supp_tab:entire_layout}
\end{table}
\begin{table}[!b]  % [b] places at bottom of page; use [btp] to give LaTeX more options
    \centering
    
    % Table 1
    \caption{Average of Maximum Weighted $F_1$-score ($\uparrow$) for \textsc{WayCoach} experiments on \textit{nonadaptive} teaching \textbf{with previous action as input.}} 
     \label{supp_tbl:table_2_nonadaptive}
     {\renewcommand\tabularxcolumn[1]{m{#1}}
     \begin{tabularx}{\textwidth}{|l|Y|Y|Y|Y|}
       
        \hline
        \textbf{Dataset-Size} & \textbf{NonAdaptive} & \textbf{Random-CrossAttn} & \textbf{Full-CrossAttn} & \textbf{KNN-CrossAttn (Ours)} \\ \hline
        60 & \small \mbox{$0.637 \pm 0.038$} & \small \mbox{$0.633 \pm 0.023$} & \small \mbox{$\textbf{0.686} \pm \textbf{0.046}$} & \small \mbox{$0.650 \pm 0.032$} \\ \hline
        600 & \small \mbox{$0.778 \pm 0.035$} & \small \mbox{$0.785 \pm 0.012$} & \small \mbox{$\textbf{0.924}\pm \textbf{0.015}$} & \small \mbox{$0.787 \pm 0.024$}  \\ \hline
        6000 & \small \mbox{$0.863 \pm 0.013$} & \small \mbox{$0.872 \pm 0.008$} & \small \mbox{$\textbf{0.988}\pm \textbf{0.006}$} & \small \mbox{$0.894 \pm 0.012$}   \\ \hline
    \end{tabularx}
    }
\end{table}
\begin{table}[!t]  % [b] places at bottom of page; use [btp] to give LaTeX more options
    \centering
    % Table 1
    \caption{Average of Maximum Weighted $F_1$-score ($\uparrow$) for \textsc{WayCoach} experiments on \textit{nonadaptive} teaching with \textbf{no previous action as input}.} 
     \label{supp_tbl:table_3_nonadaptive}
     {\renewcommand\tabularxcolumn[1]{m{#1}}
     \begin{tabularx}{\textwidth}{|l|Y|Y|Y|Y|}
       
        \hline
        \textbf{Dataset-Size} & \textbf{NonAdaptive} & \textbf{Random-CrossAttn} & \textbf{Full-CrossAttn} & \textbf{KNN-CrossAttn (Ours)} \\ \hline
        60 & \small \mbox{$\textbf{0.616} \pm \textbf{0.044}$} & \small \mbox{$0.613 \pm 0.018$} & \small \mbox{$0.614 \pm 0.025$} & \small \mbox{$0.612 \pm 0.051$} \\ \hline
        600 & \small \mbox{$0.775 \pm 0.028$} & \small \mbox{$0.761 \pm 0.021$} & \small \mbox{$\textbf{0.780} \pm \textbf{0.038}$} & \small \mbox{$0.770 \pm 0.026$}  \\ \hline
        6000 & \small \mbox{$0.866 \pm 0.015$} & \small \mbox{$0.878 \pm 0.018$} & \small \mbox{$0.882 \pm 0.013$} & \small \mbox{$\textbf{0.890} \pm \textbf{0.012}$}   \\ \hline
    \end{tabularx}
    }
\end{table}
\begin{table}[t!]
\centering
\small
\caption{Average of Maximum Weighted $F_1$-score ($\uparrow$) for time based nearest neighbor retrieval features for \textbf{KNN-CrossAttn} and \textbf{Full-CrossAttn} in the low data regime, with and without previous action as input.}
\label{supp_tab:temporal_knn}
\begin{tabular}{c|cc||cc}
    \hline
    & \multicolumn{2}{c||}{\textbf{prev\_action\_encoding = Yes}} & \multicolumn{2}{c}{\textbf{prev\_action\_encoding = No}} \\
    \cline{2-5}
    \textbf{Dataset Size} & \textbf{Full-CrossAttn} & \textbf{KNN-CrossAttn} & \textbf{Full-CrossAttn} & \textbf{KNN-CrossAttn} \\
    \hline
    100 & \mbox{$0.567 \pm 0.047$} & \mbox{$\textbf{0.603} \pm \textbf{0.012}$} & \mbox{$0.489 \pm 0.015$} & \mbox{$\textbf{0.521} \pm \textbf{0.022}$} \\
    200 & \mbox{$\textbf{0.670} \pm \textbf{0.029}$} & \mbox{$0.655 \pm 0.025$} & \mbox{$0.542 \pm 0.024$} & \mbox{$\textbf{0.549} \pm \textbf{0.019}$} \\
    400 & \mbox{$\textbf{0.846} \pm \textbf{0.018}$} & \mbox{$0.648 \pm 0.018$} & \mbox{$0.540 \pm 0.019$} & \mbox{$\textbf{0.579} \pm \textbf{0.028}$} \\
    600 & \mbox{$\textbf{0.922} \pm \textbf{0.019}$} & \mbox{$0.669 \pm 0.016$} & \mbox{$0.560 \pm 0.013$} & \mbox{$\textbf{0.617} \pm \textbf{0.028}$} \\
    \hline
\end{tabular}
\end{table}
\subsection{Robustness to noise in KNN features}

We explored the robustness of \textbf{KNN-CrossAttn} model to noise in the nearest neighbor selection process. 
For a given noise level, $\epsilon$, in order to sample $K$ neighbors for the current scenario $q_{curr}$ at time $t$, we consider a sampling set 
\[
    \mathcal{K}_{sample} = \mathcal{P} \cup \mathcal{R}
\]
\begin{wrapfigure}{r}{0.43\textwidth}
    \centering
    \vspace{-0.5cm}
    \includegraphics[width=\linewidth]{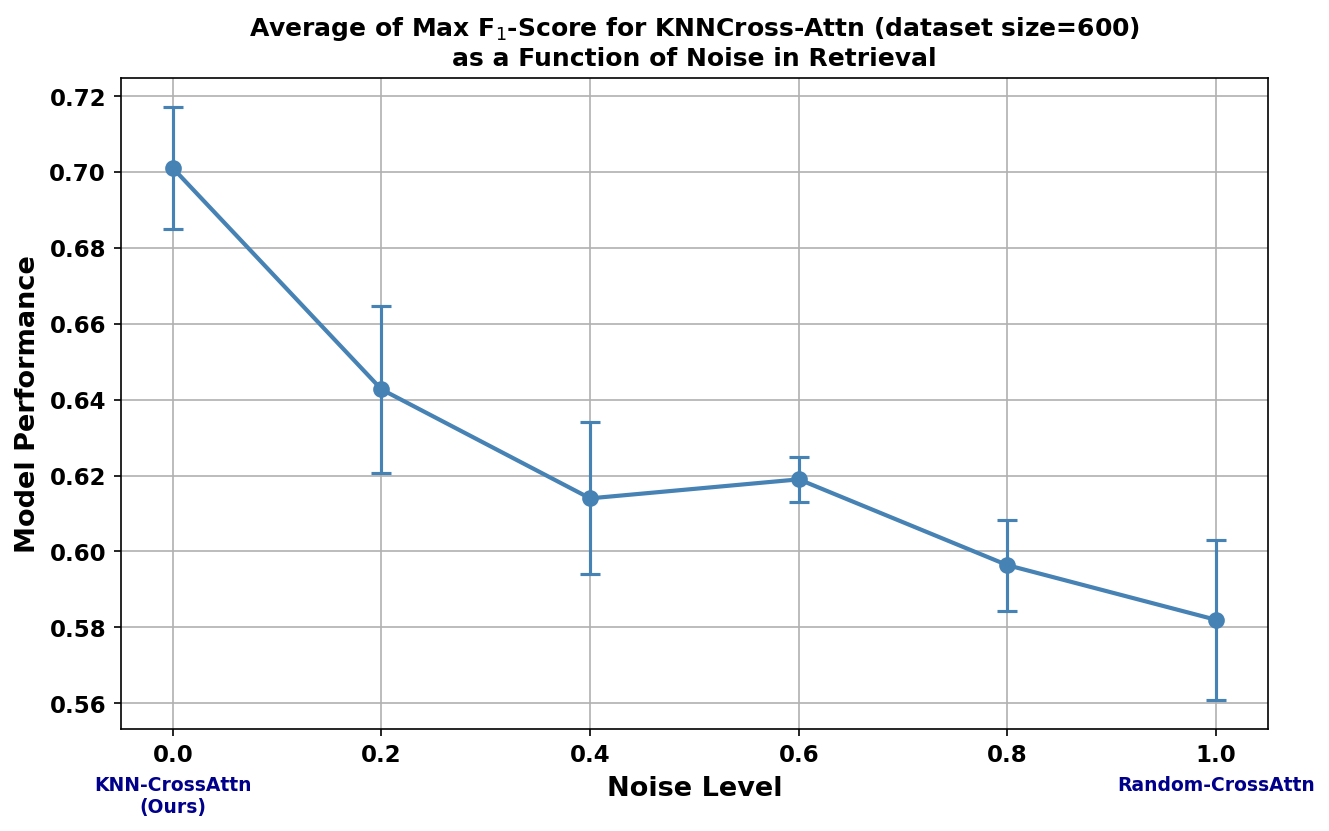}
    \caption{\textbf{KNN-CrossAttn} model performance as a function of noise in the nearest neighbor retrieval process. The degradation is higher ($\sim$$6\%$) in the beginning followed by a steady decrease ($\sim$$2$-$3\%$) all the way till \textbf{Random-CrossAttn}.}
    \label{supp_fig:noise_vs_performance}
    \vspace{-0.5cm}
\end{wrapfigure}
,where $\mathcal{R}$ is the set of the first $round(\epsilon\cdot(max(0, t- |\mathcal{P}|)))$ number of scenarios in $\mathcal{H} \setminus \mathcal{P}$ (sorted in increasing distance from $q_{curr}$). 

We then randomly sample $K$ neighbors from $\mathcal{K}_{sample}$ to form the model's past input. Note that, when  when $\epsilon=0.0$, $\mathcal{K}_{sample} = \mathcal{P}$ and matches the \textbf{KNN-CrossAttn} condition. When $\epsilon=1.0$, $\mathcal{K}_{sample}$ becomes all of the past history $\mathcal{H}$ and is equivalent to the \textbf{Random-CrossAttn} condition. 
Figure \ref{supp_fig:noise_vs_performance}, shows the change in model performance as a function of noise level. We observe that there is close to a $6\%$ drop in model performance at the beginning (still with a $F_1$-score of $65\%$) followed by a more steady decline (of around ~$2-3\%$) per increase (in steps of 0.2) in noise level. 

\subsection{Results on temporal KNN similarity features}
In this section we report results on a dataset in which the skill estimate for the adaptive teacher is computed on scenarios from the previous $|P|$ timesteps with respect to $q_{curr}$. This dataset uses a much simpler \textit{time-based} nearest neighbor feature compared to the dataset used for the results presented in Table \ref{tbl:table_1} in which the skill estimate was computed on a set of nearest scenarios computed on behavioral and geometrical features (such as road curvature and velocity). In Table \ref{supp_tab:temporal_knn}, we compare \textbf{Full-CrossAttn} and \textbf{KNN-CrossAttn} models on this dataset in the low data regime (dataset sizes ranging from 100 - 600) with and without previous action as part of the input. We observe that when previous teaching action is part of the input, the \textbf{Full-CrossAttn} model does better than \textbf{KNN-CrossAttn} even for dataset size=200, which is a lower cross-over point than the results reported in Figure ~\ref{fig:model_performance_vs_dataset_size} (which used the dataset with behavior and geometrical features), indicating that a time-based KNN is a much easier task for \textbf{Full-CrossAttn} to learn using its in-context capabilities. However, when previous action is not part of the input (Table \ref{supp_tab:temporal_knn} columns 3 and 4), \textbf{KNN-CrossAttn} achieves better performance for all dataset sizes in the low-data regimes. Overall the results indicate that the trends observed are agnostic to the choice of features that determine the selection of $K$ nearest scenarios.

\begin{wrapfigure}{r}{0.43\textwidth}
    % \vspace{-1.5cm}
    \centering
    \includegraphics[width=0.55\linewidth]{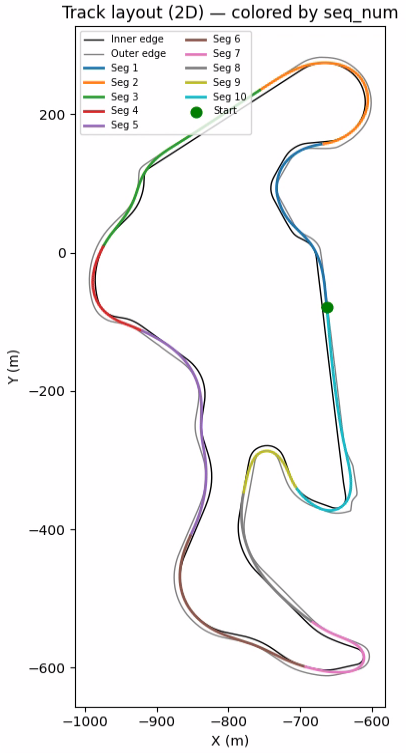}
    \caption{Map of Thunderhill West color coded according to segments.}
    \label{supp_fig:map}
    \vspace{-1cm}
\end{wrapfigure}

\subsection{Naive model mismatch results}
In this section, we present results from a naive model mismatch experiment in which we run the non-adaptive teacher's decision rule on the scenarios in a student sequence generated by the adaptive teacher. When we treat the non-adaptive teacher's actions as non-neural baseline predictions on the adaptive teacher's scenarios, the weighted $F_1$-score is 0.604 which is slightly lower than the performance of \textbf{KNN-CrossAttn} at very low data regimes (dataset-size = 60, Table \ref{tbl:table_1}, top row) and much lower (by about 10-20$\%$) for larger dataset sizes.
\begin{wrapfigure}{r}{0.43\textwidth}
    \centering
    % \vspace{-1cm}
    \includegraphics[width=\linewidth]{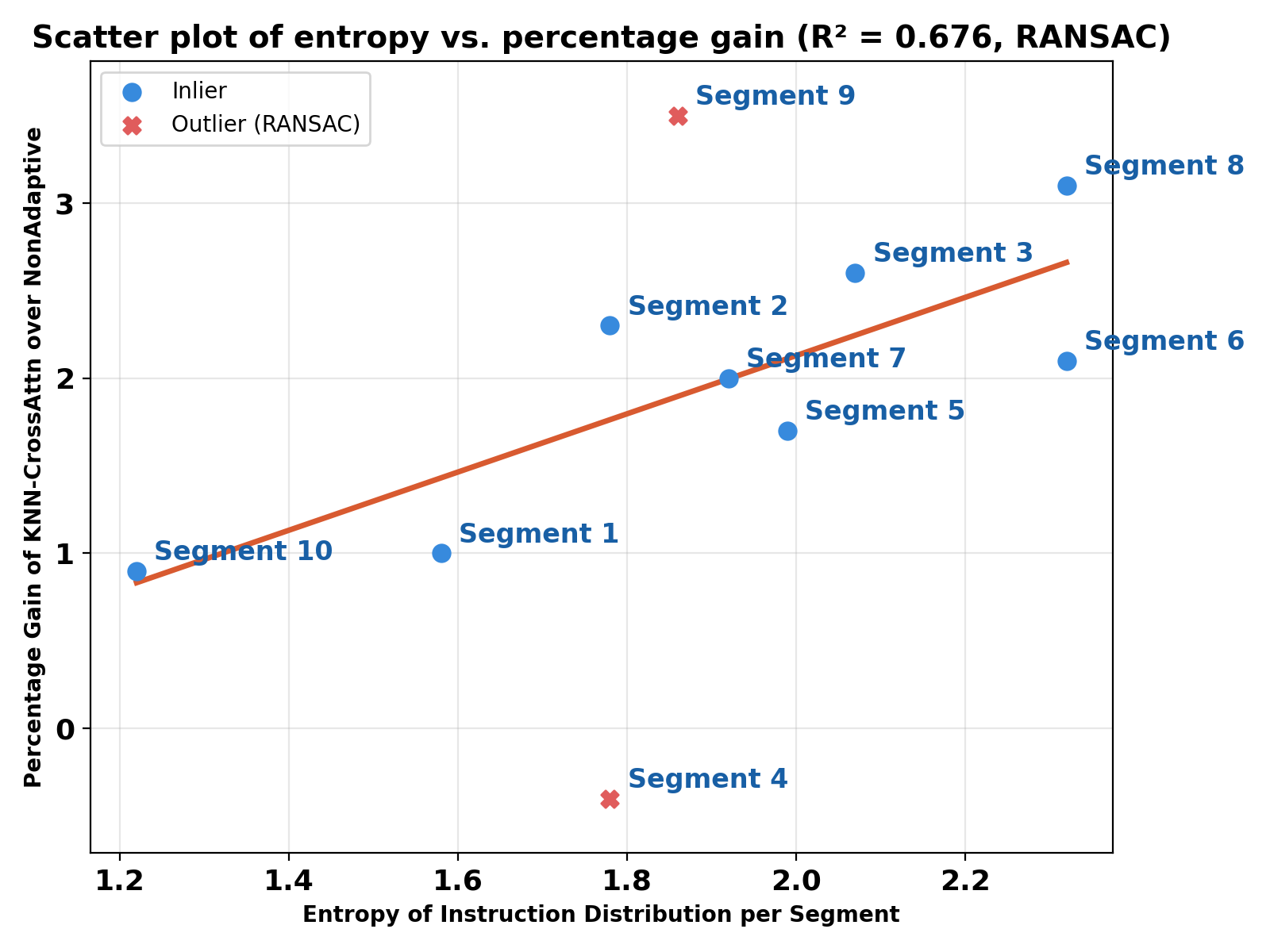}
    \caption{Percentage gain in model performance for \textbf{KNN-CrossAttn} compared to \textbf{NonAdaptive} as a function of entropy of instruction distribution per segment. Relative gain of the adaptive model is higher in those segments where entropy of instruction distribution is higher.}
    \label{supp_fig:entropy_vs_gain}
    % \vspace{-1cm}
\end{wrapfigure}
\section{Additional \textsc{SimCoachCorpus} Results}
In this section we present additional results on \textsc{SimCoachCorpus} that focus on how the adaptive model performs on each instruction category as well as on each track segment. 
\subsection{Model performance per instruction category}

% \vspace{-2.5cm}
Table \ref{supp_tbl:table_8_instruction_category} presents the results on how the \textbf{KNN-CrossAttn} model fairs against a \textbf{NonAdaptive} model for each valid instruction category. First, we observe that the \textbf{KNN-CrossAttn} model is better than \textbf{NonAdaptive} on all categories. In particular, we notice that on the category ``throttle off'', the  \textbf{KNN-CrossAttn} has \textbf{5.2$\%$} gain over the \textbf{NonAdaptive}. 

In HPDE driving, it is usually the case that the initial laps are slower (smaller throttle application) and the coach encourages the student to be more confident and apply more throttle as the session progresses. As the student gets more comfortable with applying the throttle, instructions related to when to take the foot \textit{off the throttle} also becomes relevant. Particularly, in the track (Thunderhill West) used in \textsc{SimCoachCorpus}, there are a couple of sharp turns as seen in Figure \ref{supp_fig:map} where it is critical to take the foot off the throttle pedal. \textbf{KNN-CrossAttn} is able to model these `throttle off' instructions more effectively than \textbf{NonAdaptive} because it is able to reason about how the teaching and behavior evolves over past laps. 
\begin{table}[t!]
\caption{Comparison of average of maximum $F_1$-score per instruction category on \textsc{SimCoachCorpus} between \textbf{NonAdaptive} and \textbf{KNN-CrossAttn} models. Higher gains are observed for \textbf{KNN-CrossAttn} for categories related to slowing down (`throttle off', `brake on') the car. }
\label{supp_tbl:table_8_instruction_category}
\centering
\begin{tabular}{cccc}
% \label{tbl}
\hline
\textbf{Instruction Category} & \textbf{NonAdaptive} & \textbf{KNN-CrossAttn}  & \textbf{Delta Performance $\%$}\\
\hline
stay / move right       & 0.725 $\pm$ 0.069 & \textbf{0.743 $\pm$ 0.065} & +1.8\\
stay / move left        & 0.579 $\pm$ 0.112 & \textbf{0.621 $\pm$ 0.112} & +4.2\\
throttle on             & 0.745 $\pm$ 0.047 & \textbf{0.761 $\pm$ 0.045} & +1.6\\
throttle off            & 0.711 $\pm$ 0.051 & \textbf{0.763 $\pm$ 0.036} & \textbf{+5.2}\\
throttle stay           & 0.411 $\pm$ 0.147 & \textbf{0.422 $\pm$ 0.151} & +1.1\\
brake on                & 0.782 $\pm$ 0.071 & \textbf{0.815 $\pm$ 0.024} & +3.3\\
brake off               & 0.545 $\pm$ 0.271 & \textbf{0.564 $\pm$ 0.287} & +1.9\\
\hline
\end{tabular}

\label{tab:results}
\end{table}
\begin{table}[t!]
\caption{Comparison of average of maximum $F_1$-score per track segment on \textsc{SimCoachCorpus} between \textbf{NonAdaptive} and \textbf{KNN-CrossAttn} models. }
\label{supp_tbl:table_9_per_segment}
\centering
\begin{tabular}{cccc}
% \label{tbl}
\hline
Segment ID & \textbf{NonAdaptive} & \textbf{KNN-CrossAttn} & \textbf{Delta Performance $\%$} \\
\hline
$1$       & 0.816 $\pm$ 0.063 & \textbf{0.826 $\pm$ 0.062} & +1.0 \\
$2$       & 0.702 $\pm$ 0.064 & \textbf{0.725 $\pm$ 0.051} & +2.3\\
$3$       & 0.666 $\pm$ 0.090 & \textbf{0.692 $\pm$ 0.076} & +2.6\\
$4$       & \textbf{0.700 $\pm$ 0.108} & 0.696 $\pm$ 0.117 & -0.4\\
$5$       & 0.567 $\pm$ 0.101 & \textbf{0.584 $\pm$ 0.099} & +1.7\\
$6$       & 0.658 $\pm$ 0.081 & \textbf{0.679 $\pm$ 0.086} & +2.1\\
$7$       & 0.718 $\pm$ 0.065 & 0.\textbf{738 $\pm$ 0.050} & +2.0\\
$8$       & 0.700 $\pm$ 0.107 & \textbf{0.731 $\pm$ 0.088} & +3.1\\
$9$       & 0.768 $\pm$ 0.056 & \textbf{0.803 $\pm$ 0.065} & +3.5\\
$10$      & 0.885 $\pm$ 0.013 & \textbf{0.894 $\pm$ 0.014} & +0.9\\
\hline
\end{tabular}

\label{tab:results}
\end{table}

\subsection{Model performance per track segment}
\begin{figure}[t!]
    \begin{subfigure}[b]{0.3\textwidth}
        \centering
        \includegraphics[width=\textwidth]{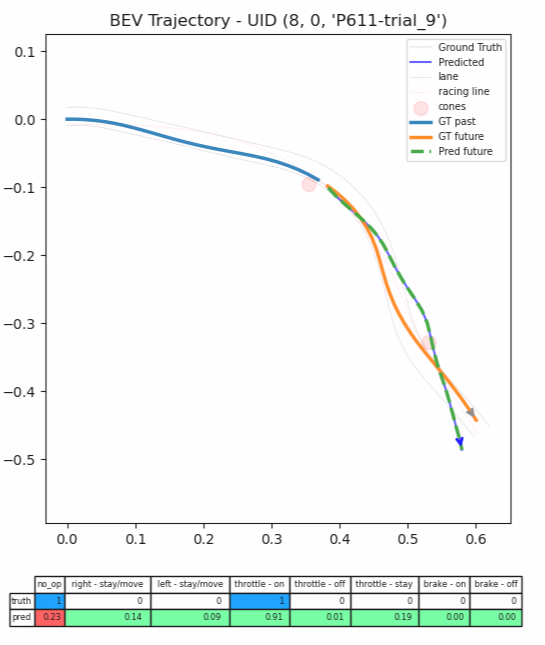}
        \caption{}
        \label{fig:traj_pred_a}
    \end{subfigure}
    \hfill
    \begin{subfigure}[b]{0.3\textwidth}
        \centering
        \includegraphics[width=\textwidth]{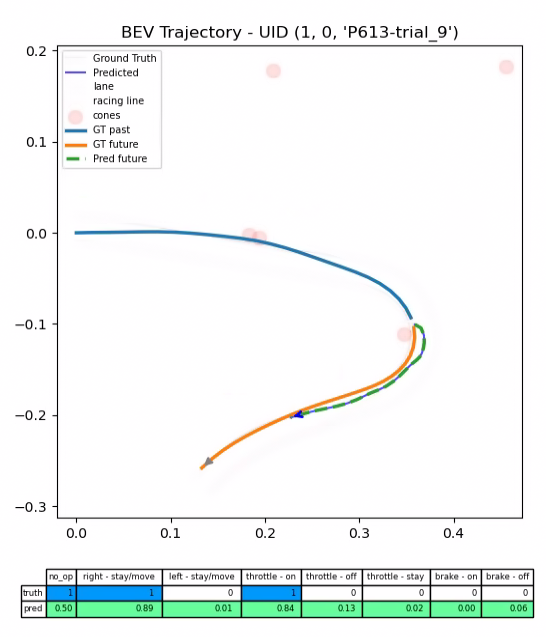}
        \caption{}
        \label{fig:traj_pred_b}
    \end{subfigure}
    \hfill
    \begin{subfigure}[b]{0.3\textwidth}
        \centering
        \includegraphics[width=\textwidth]{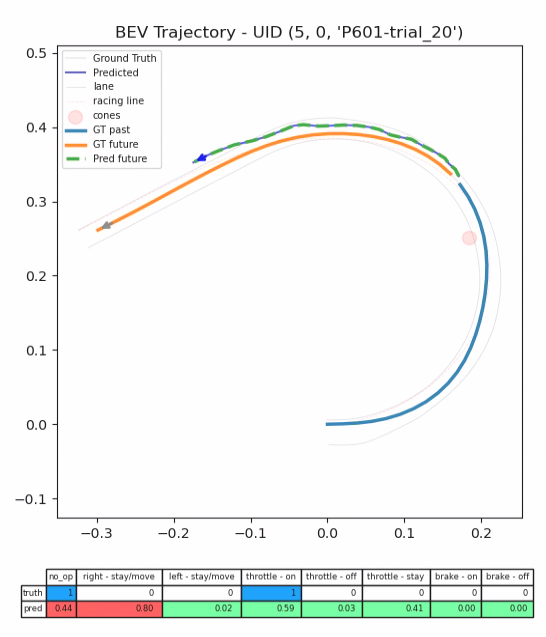}
        \caption{}
        \label{fig:traj_pred_c}
    \end{subfigure}
    \caption{Examples of trajectory and teaching action predictions from the \textbf{KNN-CrossAttn} model. Note that the trajectory prediction closely tracks the ground truth and is able to correlate behavior with map geometry. The teacher action predictions closely track the ground truth with occasional false positives that are still contextually relevant.}
    \label{supp_fig:model_outputs}
\end{figure}

Figure \ref{supp_fig:map} presents a map the track on which \textsc{SimCoachCorpus} data was collected. The map is divided into ten segments, where each segment roughly corresponds to a corner complex. In HPDE, the geometrical features of a particular segment is a significant factor that determines what coaching instructions are provided in that segment. For example, in a sharp hairpin turn the coach might \textit{always} tell the student to ``slow down'' or ``step on the brake'', whereas in other segments such as a slalom (Segment 5 or 6) the instructions could be a lot more varied (higher entropy for the instruction distribution). Table \ref{supp_tbl:table_9_per_segment} presents model performance and the relative gains between \textbf{KNN-CrossAttn} and \textbf{NonAdaptive} models and we observe that except for Segment 4, \textbf{KNN-CrossAttn} achieves better performance than the \textbf{NonAdaptive} counterpart. 
 
 Figure \ref{supp_fig:entropy_vs_gain} shows how the relative gains change as a function of the entropy of the instruction distribution per segment. A RANSAC based linear fit revealed a strong positive correlation of $R \approx 0.822$, indicating that the adaptive model is better able to fit the distribution of instruction in those segments where the entropy is higher. In particular, for Segment 10, as it is the last one before the end gate, the coach always asked the student to ``step on the throttle'' and speed through the end gate, resulting in an almost unimodal distribution. Correspondingly, the relative gain is the smallest because there is nothing to learn from reasoning about the past. All necessary information to predict ``throttle on'' is present in the geometry of the segment. 

\subsection{Visualization of model outputs}
Figure \ref{supp_fig:model_outputs} presents a few output examples from \textbf{KNN-CrossAttn} model on the \textsc{SimCoachCorpus}. We see that the trajectory predictions closely track the ground truth, indicating that the model can correlate past observed behavior and local map geometry to generate future behavior predictions. For (b) and (c) we see that although the geometry of the predicted trajectories match, the model has underestimated the speed of the future behavior and in both cases, yet the model is able to predict the ``throttle-on'' category correctly. In (b) we also observe that the predicted trajectory is to the \textit{left} of the ground truth trajectory and the model is also able to correctly predict the ``right - stay / move'' instruction category. The model outputs are not without imperfections. In (c), we see a false positive (the model predicts ``right - stay / move'' when the ground truth does not have it). However, in this particular situation the false positive is still contextually relevant because at the end of the straightaway is a left turn (end of the straight segment at the top of Figure \ref{supp_fig:map}, middle of Segment 3), where it is important to start the left turn maneuver from the right side of the track. 

\section{\textsc{WayCoach} construction details}

In this section, we present some of the details regarding the construction of \textsc{WayCoach}, with a particular focus on the design of trajectory filters, and the teacher action generation mechanism. The construction of \textsc{WayCoach} is done in multiple stages starting with scenario filtering. 
\subsection{Interactive scenario filtering}
Most of the scenarios included in Waymo Open Motion Dataset are of non-interactive driving with minimal interaction with other agents on the road. Although driving instruction is relevant in such scenarios, it is most critical when the ego-car needs to \textit{interact} with other agents where the ego-driver needs to understand when to slow down and speed up or remain the same. To this end, we extracted a set of\textit{ yield-focused }scenarios from the Waymo interactive scenario set (\texttt{validation\_interactive} folder in the full WOMD) that contain interactions between two agents. To extract yields, we crafted trajectory filters based on various heuristics. We describe these next.  
\subsection{Heuristic Filters for Yield-Scenario Detection} 
Each scenario consists of $N$ agents and we only consider the subset of vehicle agents for filtering. First, we enumerate all ordered pairs of relevant vehicle agents. 
The filtering setup used non-interpolated trajectories, 11 past steps and 80 future steps at 0.1 s resolution, and loaded up to 64 agents per scene.

For each ordered pair $(i,j)$, agent $i$ was treated as the following/merging agent and agent $j$ as the leading agent. Let $P_i$ and $P_j$ denote their 2D trajectory polylines, and let $B(P, r)$ denote a geometric buffer (a tube-like neighborhood) of radius $r$. A candidate shared yield region was defined as
\[
    \mathcal{M}_{ij} = B(P_i, 0.5) \cap B(P_j, 0.5).
\]

\begin{algorithm}[t]
\caption{Yield acceptance criterion for an ordered agent pair $(i,j)$}
\label{alg:yield-acceptance}
\begin{algorithmic}[1]
\Require Trajectories $P_i,P_j$; arrival-time maps $t_i(\cdot),t_j(\cdot)$;
         overlap set $\mathcal{M}_{ij}$; threshold $t_{\mathrm{yield}}$
\Ensure  \textbf{true} if $(i,j)$ is accepted as a yield, \textbf{false} otherwise
\Statex
\Function{IsYield}{$i,j$}
    \If{$i$ is not a relevant vehicle \textbf{ or } $j$ is not a relevant vehicle}
        \State \Return \textbf{false} \Comment{(1) both agents must be relevant vehicles}
    \EndIf
    \If{$\mathcal{M}_{ij} = \emptyset$}
        \State \Return \textbf{false} \Comment{(2) agents must come spatially close}
    \EndIf
    \If{$P_i$ has no point in $\mathcal{M}_{ij}$ \textbf{ or } $P_j$ has no point in $\mathcal{M}_{ij}$}
        \State \Return \textbf{false} \Comment{(3) both need a point inside the overlap}
    \EndIf
    \State $k_i \gets \min\{\, k : P_i(k) \in \mathcal{M}_{ij} \,\}$
        \Comment{first index of $i$ inside overlap}
    \State $k_j \gets \min\{\, k : P_j(k) \in \mathcal{M}_{ij} \,\}$
        \Comment{first index of $j$ inside overlap}
    \If{$k_i < k_j$}
        \State \Return \textbf{false} \Comment{(4) $j$ must reach the region first ($k_i \geq k_j$)}
    \EndIf
    \State $k_i^{\min} \gets \arg\min_k \lVert P_i(k+1)-P_i(k)\rVert^2$
    \If{$k_i^{\min} \geq k_i$}
        \State \Return \textbf{false} \Comment{(5) min-speed point must precede entry}
    \EndIf
    \State $P_i^{\mathrm{approach}} \gets P_i[\,0 : t_i(k_i)\,]$
    \State $P_j^{\mathrm{approach}} \gets P_j[\,0 : t_j(k_j)\,]$
    \If{$\ell(P_i^{\mathrm{approach}}) \leq 5.0$ \textbf{ or } $\ell(P_j^{\mathrm{approach}}) \leq 5.0$}
        \State \Return \textbf{false} \Comment{(6) approach paths must be long enough}
    \EndIf
    \State $\Delta t \gets t_i(k_i) - t_j(k_j)$
    \If{$\Delta t \leq 0$ \textbf{ or } $\Delta t \geq t_{\mathrm{yield}}$}
        \State \Return \textbf{false} \Comment{(7) $0 < \Delta t < t_{\mathrm{yield}}$}
    \EndIf
    \If{$d\!\left(P_j(0),\, P_i^{\mathrm{approach}}\right) \leq 5.0$}
        \State \Return \textbf{false} \Comment{(8) $j$ must start far from $i$'s approach path}
    \EndIf
    \State \Return \textbf{true}
\EndFunction
\end{algorithmic}
\end{algorithm}

\begin{wraptable}{r}{0.42\textwidth}
  \centering
  \vspace{-2ex} % pull up to align with surrounding text
  \caption{Dataset statistics after filtering.}
  \label{supp_tab:filter_results}
  \setlength{\tabcolsep}{4pt} % tighten column spacing
  \begin{tabular}{@{}lrrr@{}}
    \toprule
    Scenario Type & Train & Val & Total \\
    \midrule
    \texttt{conservative} & 780   & 184  & 964   \\
    \texttt{aggressive}   & 733   & 149  & 882   \\
    \texttt{neutral}  & 33{,}382 & 7{,}895 & 41{,}277 \\
    \bottomrule
  \end{tabular}
  \vspace{-2ex}
\end{wraptable}
Let $k_i$ and $k_j$ be the first trajectory indices at which agents $i$ and $j$ enter $\mathcal{M}_{ij}$, respectively. $\ell(\cdot)$ denotes path length and $d(\cdot,\cdot)$ denotes the shortest Euclidean distance from a point to a path. 
The algorithm for determining if a pair $(i, j)$ is accepted as a valid yield is presented in Algorithm \ref{alg:yield-acceptance}.
Each accepted interaction produced a \texttt{yield} token with
\texttt{leading\_agent}$=$$j$ and \texttt{following\_agent}$=$$i$. The token value was the temporal yield gap we used as \textit{aggressiveness} metric, partitioning the yield occurrences into three categories: (i) \texttt{conservative}, with yield gap $>4$s and yield start time $<4$s (ii) \texttt{aggressive}, with yield gap $<3$ s and yield start time $<4$ and (iii) \texttt{neutral} in which all intermediate-gap, late-yield, or non-matching scenarios were assigned to. The code implementation of this filter has been made publicly available at \url{anonymized\_due\_to\_double\_blind}.

We split the filter results into a training and a validation set with the distribution of scenarios types shown in Table \ref{supp_tab:filter_results}. The training and validation student sequences in \textsc{WayCoach} are then sampled from the scenarios from the training and validation splits respectively.

\subsection{Student Sequence}
\begin{figure*}[t!]
  \centering
  \begin{subfigure}{\textwidth}
    \centering
    \includegraphics[width=\textwidth]{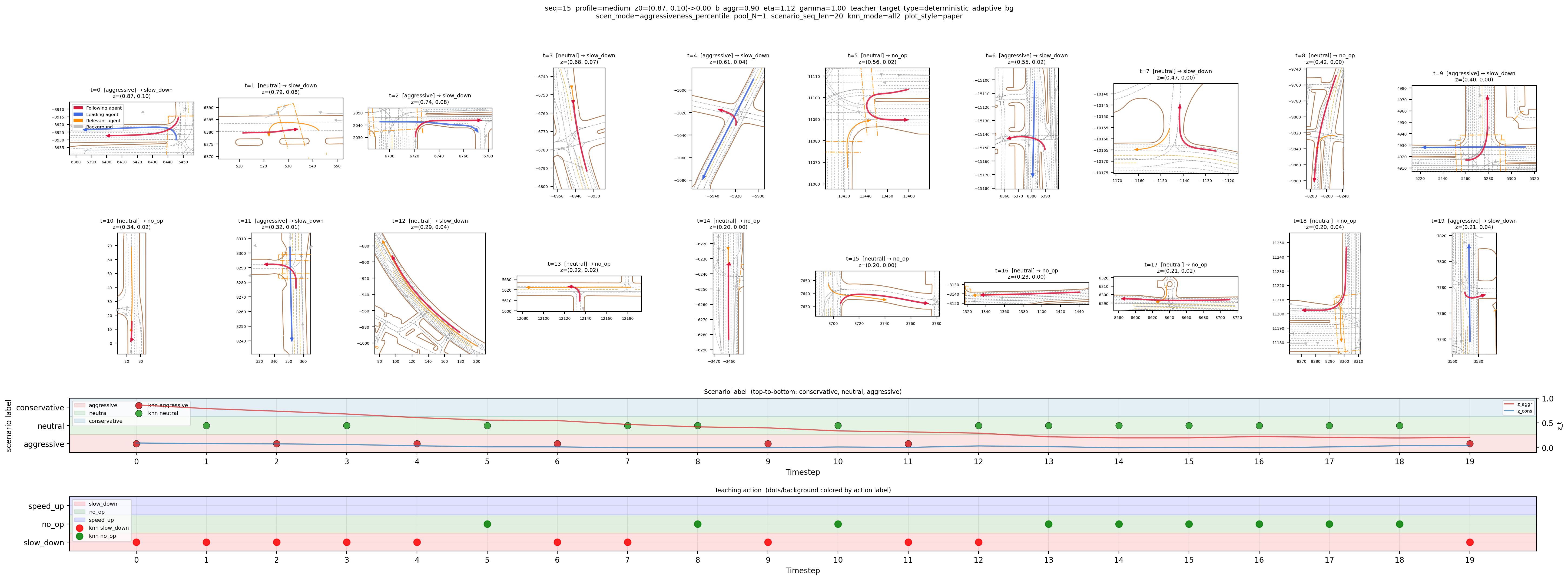}
    \caption{Student sequence from \textsc{WayCoach} with adaptive teacher — Bird's eye view.}
    \label{supp_fig:adaptive_bev}
  \end{subfigure}

  \vspace{1ex}

  \begin{subfigure}{\textwidth}
    \centering
    \includegraphics[width=\textwidth]{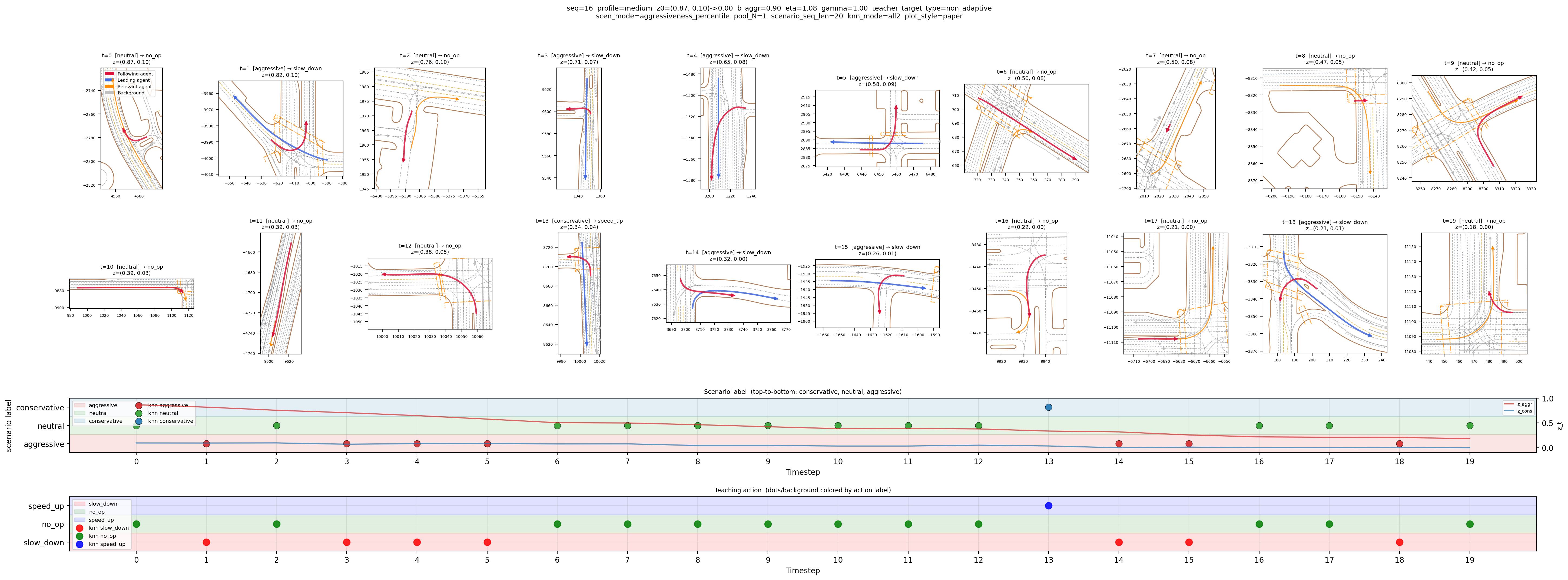}
    \caption{Student sequence from \textsc{WayCoach} with nonadaptive teacher — Bird's eye view.}
    \label{supp_fig:nonadaptive_bev}
  \end{subfigure}

  \caption{}
  \label{supp_fig:waycoach_viz}
\end{figure*}
% \vspace{-4cm}
\begin{figure*}[b!]
  \centering
  \begin{subfigure}{\textwidth}
    \centering
    \includegraphics[width=0.76\textwidth]{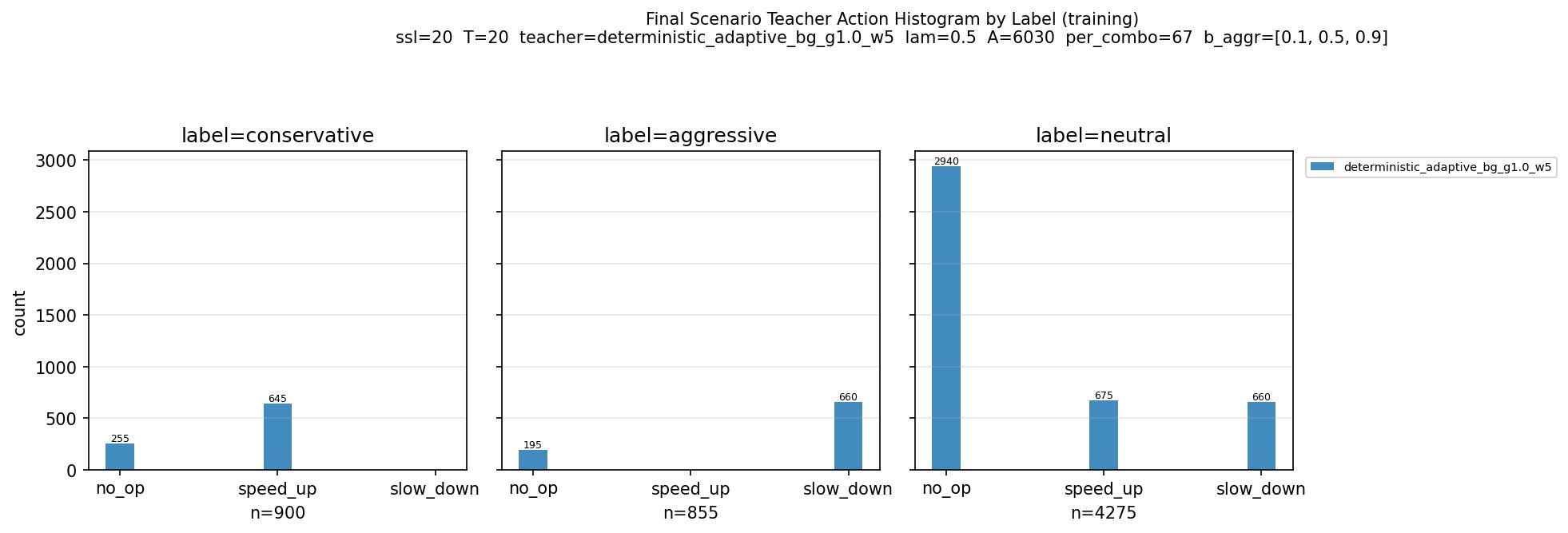}
    \caption{Adaptive teacher — instruction distribution per scenario type.}
    \label{supp_fig:adaptive_distri}
  \end{subfigure}

  \vspace{1ex}

  \begin{subfigure}{\textwidth}
    \centering
    \includegraphics[width=0.76\textwidth]{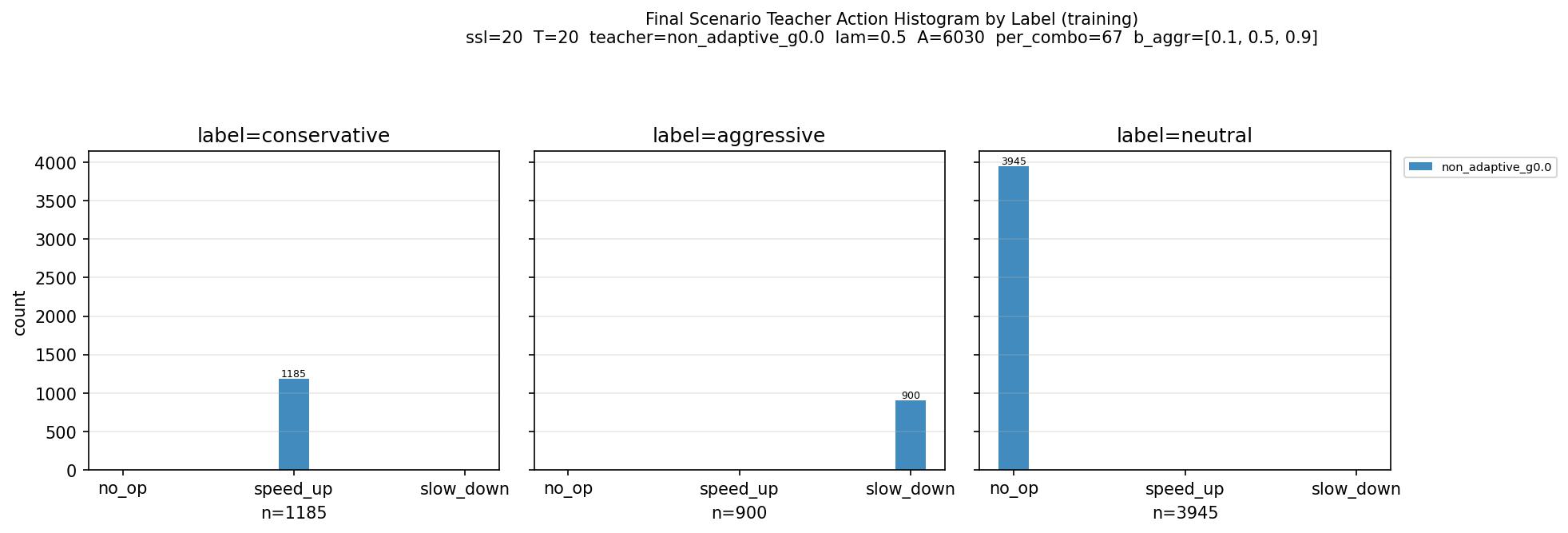}
    \caption{Nonadaptive teacher — instruction distribution per scenario type.}
    \label{supp_fig:nonadaptive_distri}
  \end{subfigure}
  \caption{}
  \label{supp_fig:waycoach_viz_distribi}
\end{figure*}
Figure \ref{supp_fig:waycoach_viz} shows a couple of examples of student sequences (along with the teacher actions and the scenario labels) from \textsc{WayCoach} dataset. Figure \ref{supp_fig:waycoach_viz} (a) and (b) are that of student sequences from an adaptive and a nonadaptive teacher respectively. In these examples, we have a learner whose $\eta$ corresponds to a \textit{medium} learning profile and $b_{aggr}$ is very high. Therefore, the scenarios sampled at the initial timesteps are primarily \texttt{aggressive} scenarios. As learning progresses, we also notice that the scenarios are primarily sampled from the \texttt{neutral} category indicating that the skill has improved over time and the student is no longer aggressive or conservative in their driving. We also observe how the teaching actions are adaptive in Figure \ref{supp_fig:waycoach_viz}(a), for example, for $t$=$3$, despite the scenario being \texttt{neutral}, the teaching action is \textit{slow down} because two out of the previous three timesteps were aggressive. This is not the case in Figure \ref{supp_fig:waycoach_viz}(b), where the mapping between scenario label and teaching action is fixed and deterministic. The exact teaching action generation algorithms for adaptive and nonadaptive teachers are provided in Algorithms \ref{alg:deterministic_adaptive_bg} and \ref{alg:non_adaptive_teacher} respectively. The non-adaptive teacher action generation algorithm follows a simple logic; if a student is \texttt{aggressive} the coaching action is \texttt{slow\_down} and if they are \texttt{conservative}, teaching action is \texttt{speed\_up}. If the student is already \texttt{neutral} in their driving, do nothing. The adaptive teacher modifies this by taking into account a skill estimate from $\mathcal{P}$. For example, if the skill estimate indicates that the student is \texttt{conservative}, then even if the current scenario is \texttt{aggressive} the teaching action will be a \texttt{no\_op}. Similarly, if the estimated skill is \texttt{aggressive}, then even if the current scenario is \texttt{conservative} the teaching action would not be \texttt{speed\_up}, but instead \texttt{no\_op}.

\begin{algorithm}[t]
\caption{Adaptive Teacher Action Generation}
\label{alg:deterministic_adaptive_bg}
\begin{algorithmic}[1]
\Require Skill vector $\boldsymbol{z}$=$(\alpha, \beta)$ $\in$ $[(0,1), (0,1)]$, scenario label $\ell$ $\in$ $\{\texttt{conservative}$, $\texttt{aggressive}$, $\texttt{neutral}\}$
\Ensure Action label $a \in \{\texttt{speed\_up}, \texttt{slow\_down}, \texttt{no\_op}\}$

\State $b \gets \max(0, 1-\alpha-\beta)$ \Comment{\texttt{neutral} probability mass}

\If{$\alpha > \beta$ \textbf{and} $\alpha \geq b$}
    \Comment{\texttt{conservative} skill dominates}
    \If{$\ell = \texttt{conservative}$}
        \State $a \gets \texttt{speed\_up}$
    \ElsIf{$\ell = \texttt{aggressive}$}
        \State $a \gets \texttt{no\_op}$
    \Else
        \State $a \gets \texttt{speed\_up}$
    \EndIf

\ElsIf{$\beta > \alpha$ \textbf{and} $\beta \geq b$}
    \Comment{\texttt{aggressive} skill dominates}
    \If{$\ell = \texttt{conservative}$}
        \State $a \gets \texttt{no\_op}$
    \ElsIf{$\ell = \texttt{aggressive}$}
        \State $a \gets \texttt{slow\_down}$
    \Else
        \State $a \gets \texttt{slow\_down}$
    \EndIf

\Else
    \Comment{\texttt{neutral} dominates, or there is a tie}
    \State $a \gets \texttt{no\_op}$
\EndIf

\State \Return $a$
\end{algorithmic}
\end{algorithm}
% \vspace{-3cm}
\begin{algorithm}[t!]
\caption{Non-Adaptive Teacher Action Generation}
\label{alg:non_adaptive_teacher}
\begin{algorithmic}[1]
\Require label $\ell \in \{\texttt{conservative}, \texttt{aggressive}, \texttt{neutral}\}$
\Ensure Action label $a \in \{\texttt{speed\_up}, \texttt{slow\_down}, \texttt{no\_op}\}$
\If{$\ell = \texttt{aggressive}$}
    \Comment{if driver is \texttt{aggressive}, slow down}
    \State $a \gets \texttt{slow\_down}$
\ElsIf{$\ell = \texttt{conservative}$}
    \Comment{if driver is \texttt{conservative}, speed up}
    \State $a \gets \texttt{speed\_up}$
\ElsIf{$\ell = \texttt{neutral}$}
    \Comment{if driver is \texttt{neutral}, do nothing}
    \State $a \gets \texttt{no\_op}$
\EndIf
% \EndIf

\State \Return $a$
\end{algorithmic}
\end{algorithm}

Figure \ref{supp_fig:waycoach_viz_distribi}(a) and (b) shows the conditional distribution of teaching actions given the scenario labels for adaptive and nonadaptive teachers. In Figure \ref{supp_fig:waycoach_viz_distribi}(a) we observe a clear bimodality in the distribution of actions for each scenario label. In contrast, the conditional distribution of teaching actions for the \textit{nonadaptive} teacher (Figure \ref{supp_fig:waycoach_viz} (b)) is unimodal as it follows a simple decision rule which deterministically maps the scenario to a single teaching action. 

\end{document}